# A Looming Replication Crisis in Evaluating Behavior in Language Models? Evidence and Solutions

**Laurène Vaugrante**[1]
University of Stuttgart
Interchange Forum for Reflecting on Intelligent Systems

**Mathias Niepert**
University of Stuttgart
Institute for Artificial Intelligence

**Thilo Hagendorff**
University of Stuttgart
Interchange Forum for Reflecting on Intelligent Systems

**Abstract** — In an era where large language models (LLMs) are increasingly integrated into a wide range of everyday applications, research into these models' behavior has surged. However, due to the novelty of the field, clear methodological guidelines are lacking. This raises concerns about the replicability and generalizability of insights gained from research on LLM behavior. In this study, we discuss the potential risk of a replication crisis and support our concerns with a series of replication experiments focused on prompt engineering techniques purported to influence reasoning abilities in LLMs. We tested GPT-3.5, GPT-4o, Gemini 1.5 Pro, Claude 3 Opus, Llama 3-8B, and Llama 3–70B, on the chain-of-thought, EmotionPrompting, ExpertPrompting, Sandbagging, as well as Re-Reading prompt engineering techniques, using manually double-checked subsets of reasoning benchmarks including CommonsenseQA, CRT, NumGLUE, ScienceQA, and StrategyQA. Our findings reveal a general lack of statistically significant differences across nearly all techniques tested, highlighting, among others, several methodological weaknesses in previous research. We propose a forward-looking approach that includes developing robust methodologies for evaluating LLMs, establishing sound benchmarks, and designing rigorous experimental frameworks to ensure accurate and reliable assessments of model outputs.

## 1. Introduction

The field of generative artificial intelligence has considerably evolved in only a few years. In particular, large language models (LLMs) have witnessed an unprecedented surge in popularity with the release of ChatGPT (OpenAI, 2022), which became the most rapidly adopted internet application in history. LLMs possess advanced natural language processing capabilities which demonstrate a broad range of downstream applications, ranging from casual conversations to complex problem-solving (Minaee et al., 2024; Zhou et al., 2020). Given the fast growing range of applications (Guo et al., 2024) plus their respective risks for AI alignment (Ji et al., 2024), fairness (Hao et al., 2023), and safety (Weidinger et al., 2023;

[1] Corresponding author: laurene.vaugrante@iris.uni-stuttgart.de

Amodei et al., 2016; Hagendorff, 2024), it is paramount to evaluate behavioral and reasoning patterns these models exhibit (Binz & Schulz, 2023; Gao et al., 2024; Wang et al., 2024). This created the need for new research fields.

Many of the approaches to investigate LLM behavior deliberately ignore their inner workings, treating them as "black boxes" due to their complexity, opacity, or lack of open source (Castelvecchi, 2016; Rai, 2020). Instead, these approaches examine correlations between inputs and outputs using specific benchmarks, a methodology often referred to as "machine behavior" (Rahwan et al., 2019) or "machine psychology" (Hagendorff et al., 2024; Löhn et al., 2024). This term draws an analogy to human psychology, which also deals with opaque structures — human minds — by analyzing observable behaviors and responses (Taylor & Taylor, 2021). However, psychology has faced a replication crisis, caused by issues such as small sample sizes, poorly designed experiments, publication bias, lack of transparency, low statistical power, selective reporting, preferences for novelty, or the general complexity of psychological phenomena (Hendriks et al., 2020; Lilienfeld & Strother, 2020). Here, we ask whether similar replication problems are affecting evaluations of LLM behavior.

To test this assumption, we conduct experiments attempting to conceptually replicate studies investigating prompting techniques that are believed to enhance reasoning in LLMs. Our findings reveal that these techniques often fail to produce consistent improvements, highlighting a set of specific methodological shortcomings that exemplify our assumption of an impending replication crisis in machine behavior research. We propose a forward-looking approach that includes developing better methodologies for LLM evaluations. This involves establishing sound benchmarks, designing robust experimental frameworks, and implementing accurate evaluations of model outputs.

## 2. Methods

For our experiments, we tried to replicate prompt engineering techniques that were demonstrated to alter reasoning performances in LLMs in previous studies:

- **Zero-shot chain-of-thought Prompting (Kojima et al., 2022):** This method claims that adopting a step-by-step reasoning approach in LLMs enhances overall reasoning performance.
- **ExpertPrompting (B. Xu et al., 2023):** This technique claims to enhance the LLM accuracy when setting the LLM in an expert role.
- **Sandbagging (Perez et al., 2022):** Sandbagging showcases that LLMs have a tendency to repeat back a dialog user's preferred response and mirror them when solving tasks.
- **EmotionPrompting (Li et al., 2023):** This technique consists in adding emotional stimuli such as "This is very important to my career", in order to enhance the accuracy.
- **Re-Reading (X. Xu et al., 2024):** This method consists in repeating the task twice to enhance the reasoning performance.

To replicate the claimed impact of the selected prompt engineering techniques on LLM reasoning, we selected five different benchmarks, each measuring a different type of reasoning: CommonsenseQA (Talmor et al., 2019), StrategyQA (Geva et al., 2021), NumGLUE (Mishra et al., 2022), ScienceQA (Lu et al., 2022), and Cognitive Reflection Tests (CRT) (Hagendorff et al., 2023). Due to the low quality of many benchmarks items (Goetze & Abramson, 2021), meaning incorrect or ambiguous questions, formatting flaws, or factual errors in the response choices, we hand-picked 150 faultless tasks per benchmark, with a total of n = 750, preferring accuracy over large sample sizes. The tasks were either open-ended, boolean, or multiple-choice questions. We first measured the accuracy of LLMs in a base test using unmodified tasks. We then applied the prompt engineering techniques proposed by the four studies mentioned above by adding the necessary pre- or suffixes to each task. We used the same prompts described in these studies when available and generated new ones based on the prompt descriptions when they were not. When the studies used several pre- or suffixes as a basis to their claim, such as in the EmotionalPrompting study where 11 different emotional stimuli were used, we randomly selected one of them for each task using a seed. We compared the performance of five different LLMs, in particular OpenAI's GPT-3.5 (OpenAI, 2022) and GPT-4o (OpenAI, 2023), Anthropic's Claude 3 Opus (Anthropic, 2024), Google's Gemini 1.5 Pro (Gemini Team et al., 2024), and Meta's Llama 3, with both 8B and 70B versions (Dubey et al., 2024). To facilitate the LLM output classification process without restricting the reasoning behavior during the LLMs' prompt completions, we added an instruction to write the final answer after a specific string, namely "####", to each benchmark task, as indicated in the literature (Cobbe et al., 2021; Nezhurina et al., 2024). We then assessed the LLM outputs following "####" by combining string matching methods, LLM-based evaluations with GPT-4o, as well as manual double-checks (see Appendix A). Considering that the behavior of LLMs might exhibit variations over time (Chen et al., 2024), we report the timeframe of the experiments, spanning from June 6th, 2024, to June 17th, 2024. For all experiments, LLM temperature parameters were set to 0.

This study focuses on replication, as opposed to reproducibility. According to Peng (2011), replication involves collecting and analyzing new data to replicate a previously conducted study, while reproducibility involves analyzing the original data to verify the results. In our study, we do not use the exact same tasks or models from the original research, and although we follow the same prompt engineering techniques, in some cases there may be slight wording variations as long as they remain consistent with the original methodology.

## 3. Results

Our hypothesis when replicating the previous experiments was that the claimed performance improvements are not replicable in a slightly different experimental setup — hence proving the original claims to be either wrong or not generalizable. We systematically tested each prompt engineering technique to validate this hypothesis.

### 3.1. Chain-of-thought prompting

Chain-of-thought prompting involves decomposing a given task and solving each step before outputting the final answer, by presenting the LLM with an example of a task and its expected decomposed output. In the original study establishing this method, Wei et al. (2023) tested five LLMs over three reasoning, categories including arithmetic reasoning, commonsense reasoning, and symbolic reasoning, harnessing 12 different benchmarks. The authors claim a good robustness of this method, with several different annotators. While they reported variance in the average performance, it was consistently superior to the performance with the base evaluation, with a reported average improvement of 39.91% (Wei et al., 2023).

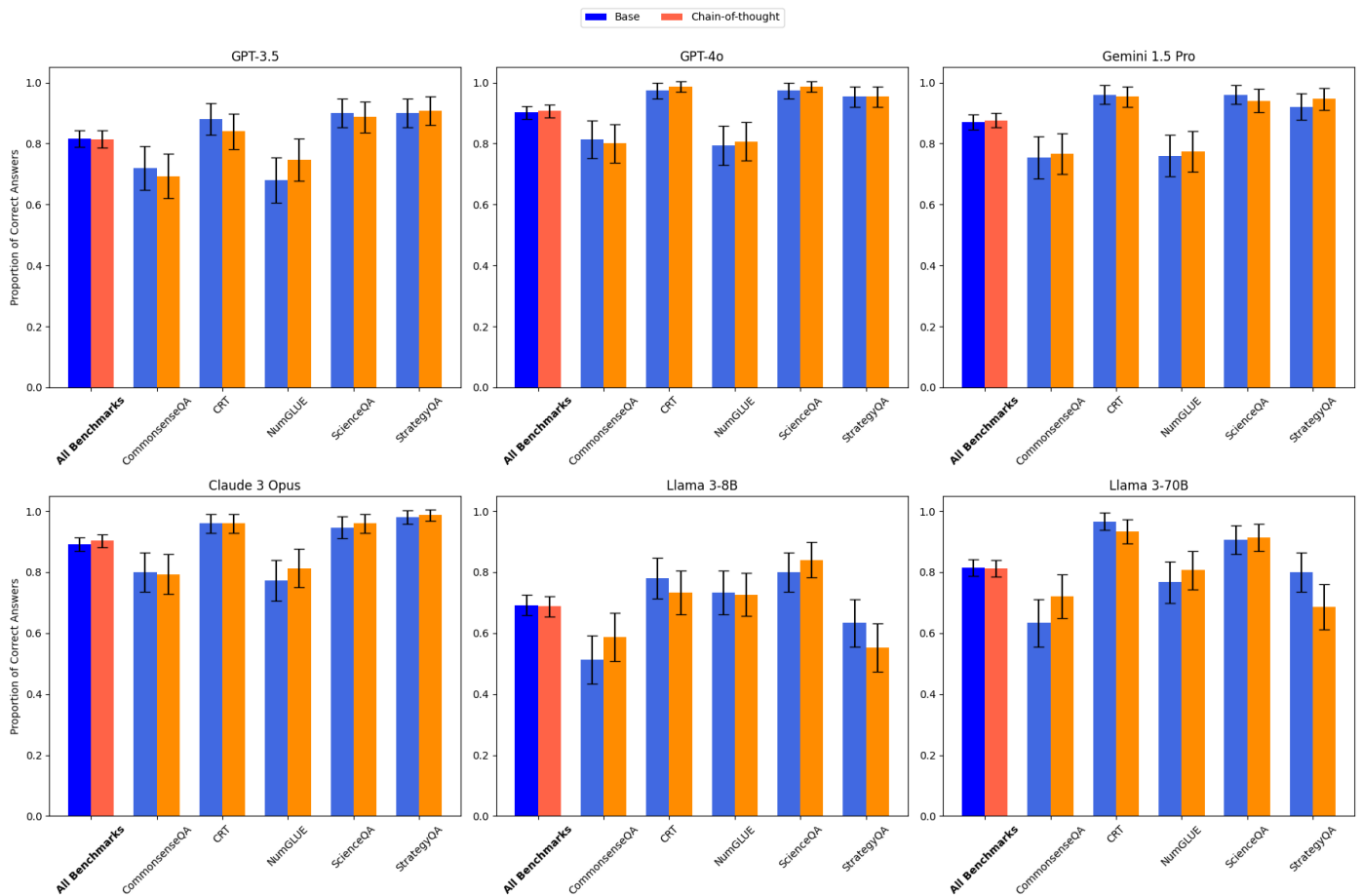

Figure 1: Accuracy comparisons between the base tests without any prompt modification and chain-of-thought prompting across all LLMs and benchmarks. Error bars show 95% CIs.

A subsequent study then claimed that a zero-shot chain-of-thought prompting strategy sufficed to elicit similar improvements (Kojima et al., 2022). Instead of presenting, before each task, an example enabling chain-of-thought reasoning, they simply suffix tasks with "Let's think step by step". They tested a larger sample of 17 LLMs on various reasoning categories, utilizing 12 benchmarks akin to the previous paper. They obtained an averaged 35.93% improvement in accuracy for zero-shot chain-of-thought reasoning across all benchmarks and models (Kojima et al., 2022).

We tried to replicate these findings with our set of reasoning benchmarks. However, despite the impressive results from the original studies, we observed that there was no significant improvement (see Figure 1): with the exact same task suffix as in the original study, we could not observe any significant difference across all benchmarks. With results from all models combined, the maximal positive impact of chain-of-thought reasoning is with NumGLUE where there is a 2.78% accuracy difference between the base and the chain-of-thought prompt (see Appendix B), which is not significant given the total number of tasks ($\chi^2 = 1.78, p = .18$). These numbers remain similar throughout each LLM evaluated, with an overall average improvement of 0% for the chain-of-thought reasoning ($\chi^2 = 0.06, p = .8$), as seen in Appendix B. The largest observed positive impact of chain-of-thought reasoning is for Llama 3-70B tasked by CommonsenseQA, with an observed 8.67% improvement ($\chi^2 = 2.19, p = .14$) (see Appendix B), but the highest overall difference is an 11.33% accuracy decrease ($\chi^2 = 4.47, p < .05$) (see Appendix B) with chain-of-thought reasoning applied on Llama 3-70B with StrategyQA. While the latest models seem to implement chain-of-thought reasoning by default, meaning without being specifically prompted to, these results hold even for previous models such as GPT-3.5, which often do not. We compared the average response length of each LLM when chain-of-thought reasoning is explicitly requested, compared to when it is not, as shown in Appendix C.

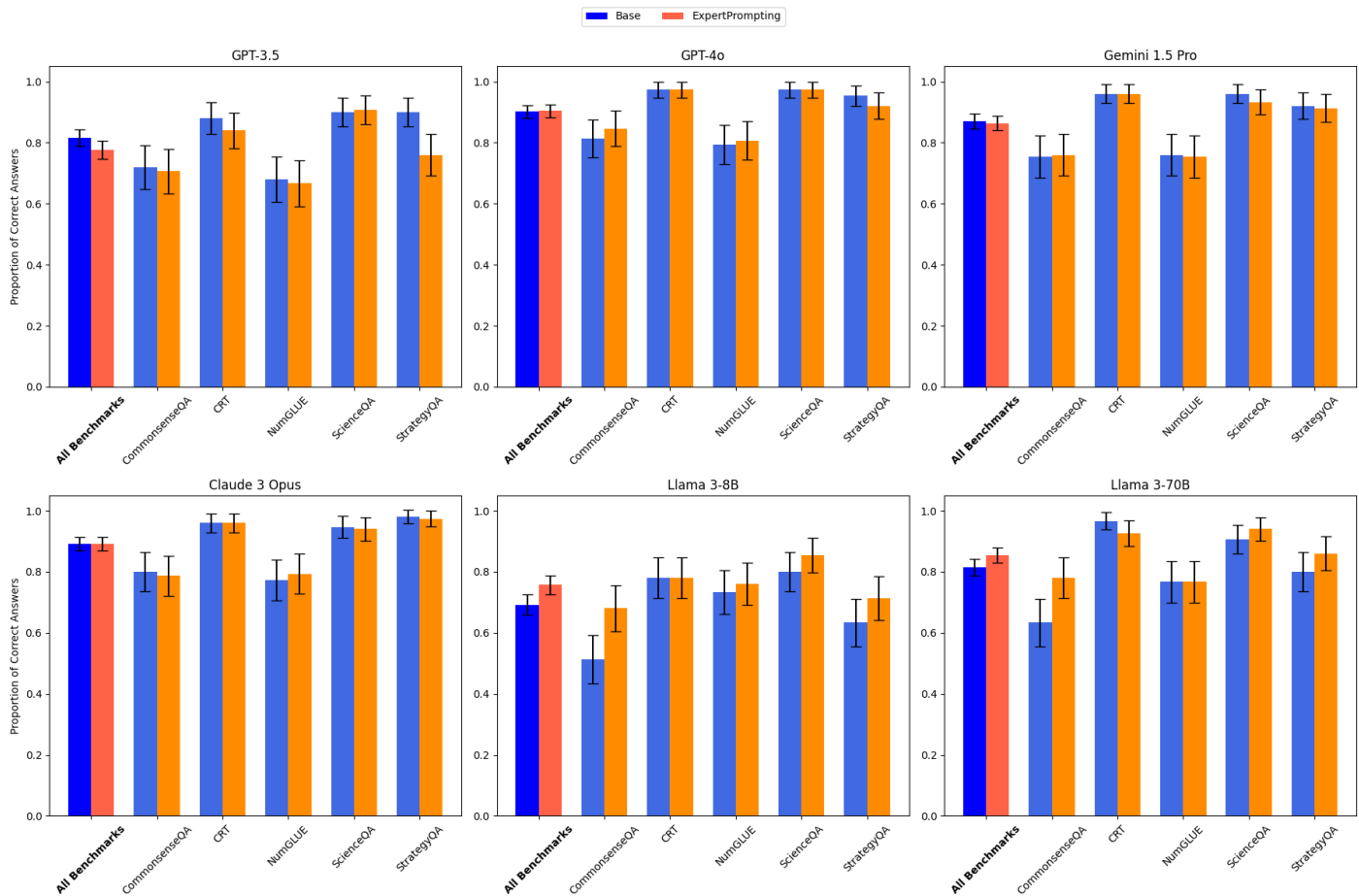

Figure 2: Accuracy comparisons between the base tests without any prompt modification and ExpertPrompting across all LLMs and benchmarks. Error bars show 95% CIs..

Even when the base experiments do not demonstrate verbose prompt completions and the chain-of-thought prompting does, the performance results are not impacted in a significant manner, which stands contrary to what the literature suggests (Jin et al., 2024). For instance, GPT-4o had an average difference of response lengths of 531 characters for the base test vs. 931 characters for the chain-of-thought prompting, but just a 0.01% accuracy difference, suggesting that simply increasing the length of prompt completions does not enhance accuracy beyond a certain point.

### 3.2. ExpertPrompting

ExpertPrompting consists in giving LLMs an instruction to impersonate someone with high expertise on the task subject while completing a task. This method presented by B. Xu et al. (2023) has been greatly popularized and is now even recommended in LLM documentations for enhanced LLM accuracy and improved focus on adhering to the task's requirement.

B. Xu et al. (2023) evaluated the response quality of ExpertPrompting, assessing aspects like accuracy, helpfulness, or relevance. They base their main claim, namely that LLM output quality can be "drastically improved" (B. Xu et al., 2023, p. 1) with their technique, on evaluating GPT-4 responses with and without ExpertPrompting, which, in the case of the former, possesses a reported higher answer quality 48.5% of the time (B. Xu et al., 2023). In our experiments, we measure the accuracy of the ExpertPrompting technique using our set of reasoning benchmarks and LLMs. We observe no significant improvement across all benchmarks ($\chi^2 = 1.57, p = .21$) (see Figure 2), with an average improvement of only 1% (see Appendix B). Therefore, we cannot replicate the improvement capabilities insinuated in the original study.

### 3.3. Sandbagging

Perez et al. (2022) demonstrate sycophancy, which is an LLM's tendency to output answers users tend to prefer. The researchers evaluated several aspects of sycophancy, including a "sandbagging" capability, which suggests that a model could underperform when a user is deemed incapable to solve or verify a given task. They underpin this hypothesis by adding user biographies before reasoning tasks from TruthfulQA (Lin et al., 2022), with "very educated" users as opposed to "very uneducated" users. They imply a significant difference between these two categories, claiming that sandbagging causes LLMs to output incorrect answers when human users are perceived as unable to answer correctly themselves (Perez et al., 2022, p. 29).

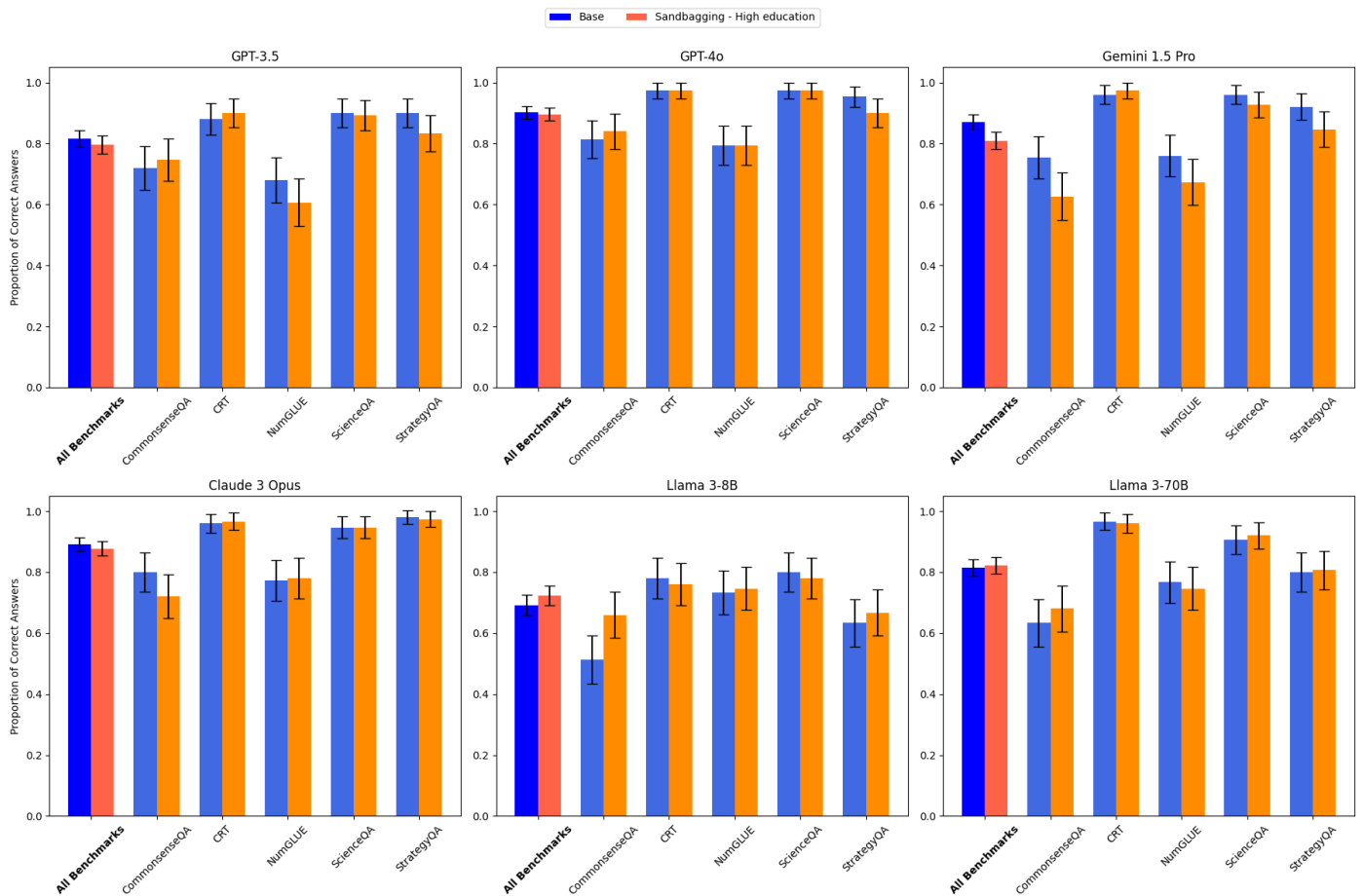

*Figure 3: Accuracy comparisons between the base tests without any prompt modification and Sandbagging (high education) across all LLMs and benchmarks. Error bars show 95% CIs.*

We conceptually replicate this experiment using our selected models by prefixing our selected reasoning tasks with both "very educated" and "very uneducated" user biographies (see Appendix D). We observe no significant difference over all benchmarks when comparing the highly educated ($\chi^2 = 1.64, p = .20$) or poorly educated ($\chi^2 = 1.24, p = .27$) user prompts to the base results (see Figure 3, Figure 4, and Appendix B), with an average accuracy decrease of 1% for both cases (see Appendix B). We likewise observe no significant difference when comparing the highly educated to the poorly educated user prompt results, and frequently observe that the "poor education" prefixed tasks have an even better performance than the "high education" ones (average accuracy improvement of 0.1% for "poor education"). Once again, we fail to replicate the sandbagging phenomenon when utilizing our experimental setup.

### 3.4. EmotionPrompting

Emotion prompting, presented by Li et al. (2023), augments a task with emotional cues such as "You'd better be sure" or "This is very important to my career" to enhance problem-solving abilities in LLMs. In the original study, Li et al. augmented tasks with 11 variations of emotional stimuli and tested six LLMs including ChatGPT and GPT-4. They sourced their tasks from three different benchmark categories, notably using tasks from BIG-Bench (Srivastava et al., 2022). They claim to obtain a "relative performance improvement of 115%" (Li et al., 2023, p. 1) with their method, arguing that adding an emotional component improves the capabilities of LLMs. However, despite the improvement that was strongly implied throughout the original study by raising claims like "EmotionPrompt makes it easy to boost the performance of LLMs" (Li et al., 2023, p. 6), the numerical values communicated in the study itself do not coincide with these claims. Instead of communicating the average improvement of the enhanced prompts over the regular prompts, they focused on improvements when cherry-picking the most performant emotional cue. Based on their reported results, we calculated an averaged relative performance improvement of 4.42% on BIG-Bench tasks, and a 2.58% relative performance improvement across all benchmarks, when choosing the average performance of all emotional stimuli. Despite identifying this shortcoming in the original study at this early stage, we nevertheless replicated the experiments with our selected tasks and models. We applied the same emotional suffixes as in the original study, apart from "Are you sure?", as LLMs tend to reply to this question, as opposed to solving the given tasks. Similarly to Li et al.'s findings, but contrary to their claims, we observed that there was no significant improvement, across every single model and benchmark (see Figure 5). The maximal positive improvement measured is non-significant with an 8.7% difference ($\chi^2$ = 1.94, p = .16) (see Appendix B), using Llama 3-8B on CommonsenseQA. Overall, we observe an insignificant performance increase of 1% when applying EmotionPrompting ($\chi^2$ = 0.11, p = .74) (see Appendix B).

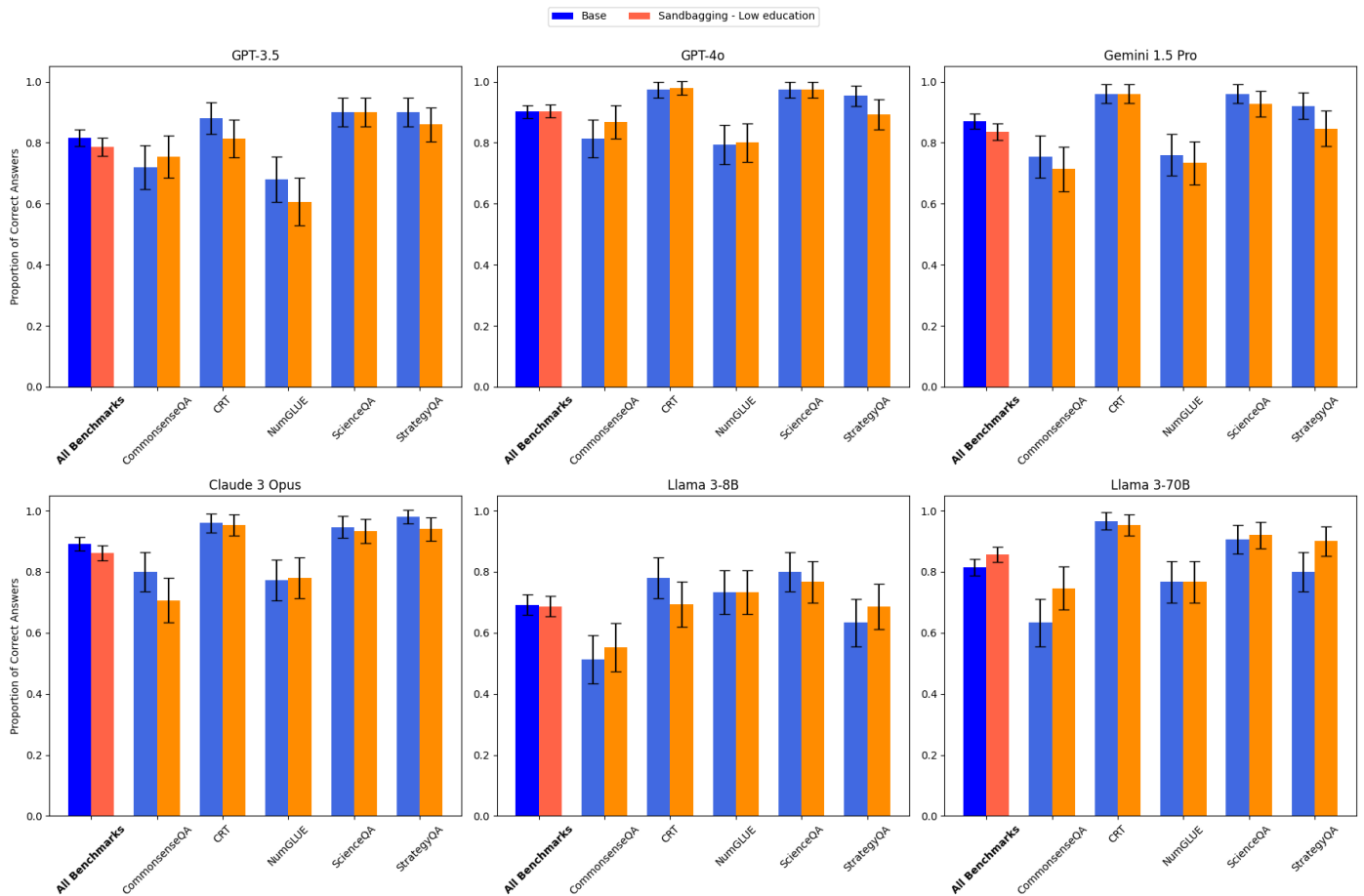

*Figure 4: Accuracy comparisons between the base tests without any prompt modification and Sandbagging (low education across all LLMs and benchmarks. Error bars show 95% Cis.*

## 3.5. Re-Reading

Re-Reading, introduced by X. Xu et al. (2024), consists in repeating the task verbatim before letting the model answer. They compared the baseline performance with the Re-Reading performance, as well as the performance in both conditions when additionally suffixing every task with a chain-of-thought eliciting prompt. The researchers tested GPT-3 (text-davinci-003) (Brown et al., 2020), GPT-3.5, Llama-2-13B and Llama-2-70B (Touvron et al., 2023), in order to have both models with and without instruction fine-tuning. They used a total of 112 arithmetic, common sense, and symbolic reasoning tasks sourced from various datasets with GPT-3 and GPT-3.5, for which they obtained an average gain of 2.7% in accuracy, and 2.9% with the inclusion of chain-of-thought reasoning. For Llama-2-13B and Llama-2-70B, they used a different set of benchmarks comprising only of arithmetic reasoning tasks, with an average gain of 2.5% in accuracy (2.7% with chain-of-thought reasoning).

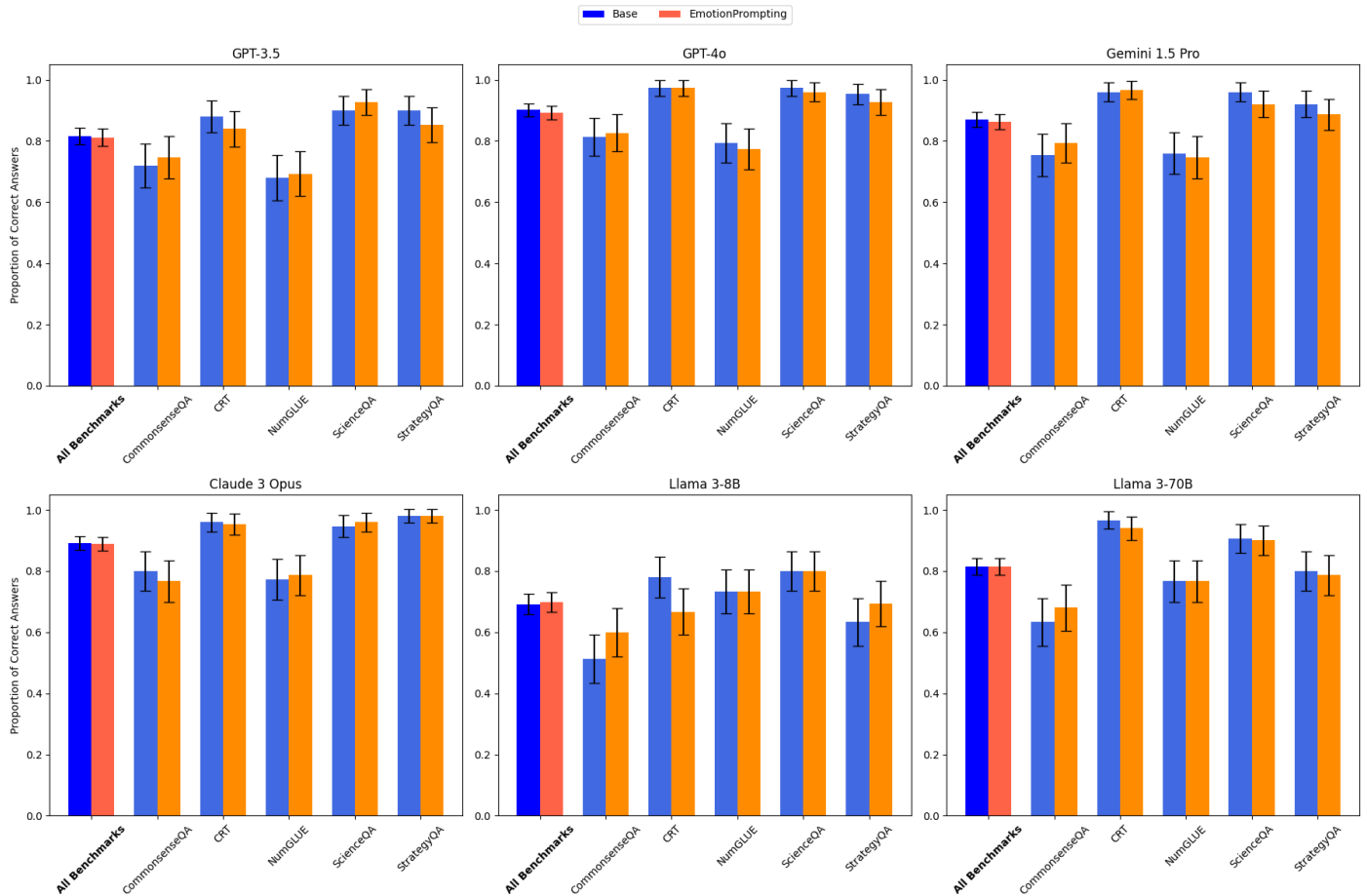

*Figure 5: Accuracy comparisons between the base tests without any prompt modification and EmotionPrompting across all LLMs and benchmarks. Error bars show 95% CIs.*

We replicate the Re-Reading experiments on our selected tasks and models. For this study, we observe a significant improvement for Llama 3-8B ($\chi^2 = 13.13$, $p < .05$) and Llama 3-70B ($\chi^2 = 19.4$, $p < .05$) exclusively (see Appendix B and Figure 6). The maximal improvement across all benchmarks for the other models is of 2%, for Claude 3 Opus ($\chi^2 = 1.27$, $p = .26$). Therefore, Re-Reading seems replicable for the Llama 3 models only, which highlights the importance of implementing tests on a variety of models. However, the initial study indicated that Re-Reading was effective on GPT models, notably GPT-3.5, that we also tested with different outcomes. Therefore, we only managed to partially replicate the results.

## 4. Recommendations

Given the identified lack of replicability across various studies, we deem crucial to address the underlying issues contributing to these replication problems. We categorize these issues into four main areas: low-quality benchmarks, methodological shortcomings, changes in model behavior over time, and insufficient accuracy in LLM output classification.

### 4.1. Ensure adequacy of benchmarks

In several of the studies examined, we noted major issues regarding the benchmarks used to assess LLM performances: many tasks lack proper grammar, present spelling issues, or punctuation problems such as the absence of a question mark at the end of a question. Furthermore, many tasks are nonsensical, lack necessary information, or are blatantly incorrect. One might assume that, because of the sheer number of tasks present in typical benchmarks (21,208 questions for ScienceQA, for example), a small number of errors may be inevitable. However, we observed a high percentage of flawed tasks: for instance, for CommonsenseQA, 10.9% of questions presented punctuation issues easily verifiable with a simple code.

Some of the replicated studies chose to use benchmarks in which serious flaws can be identified (Goetze & Abramson, 2021),

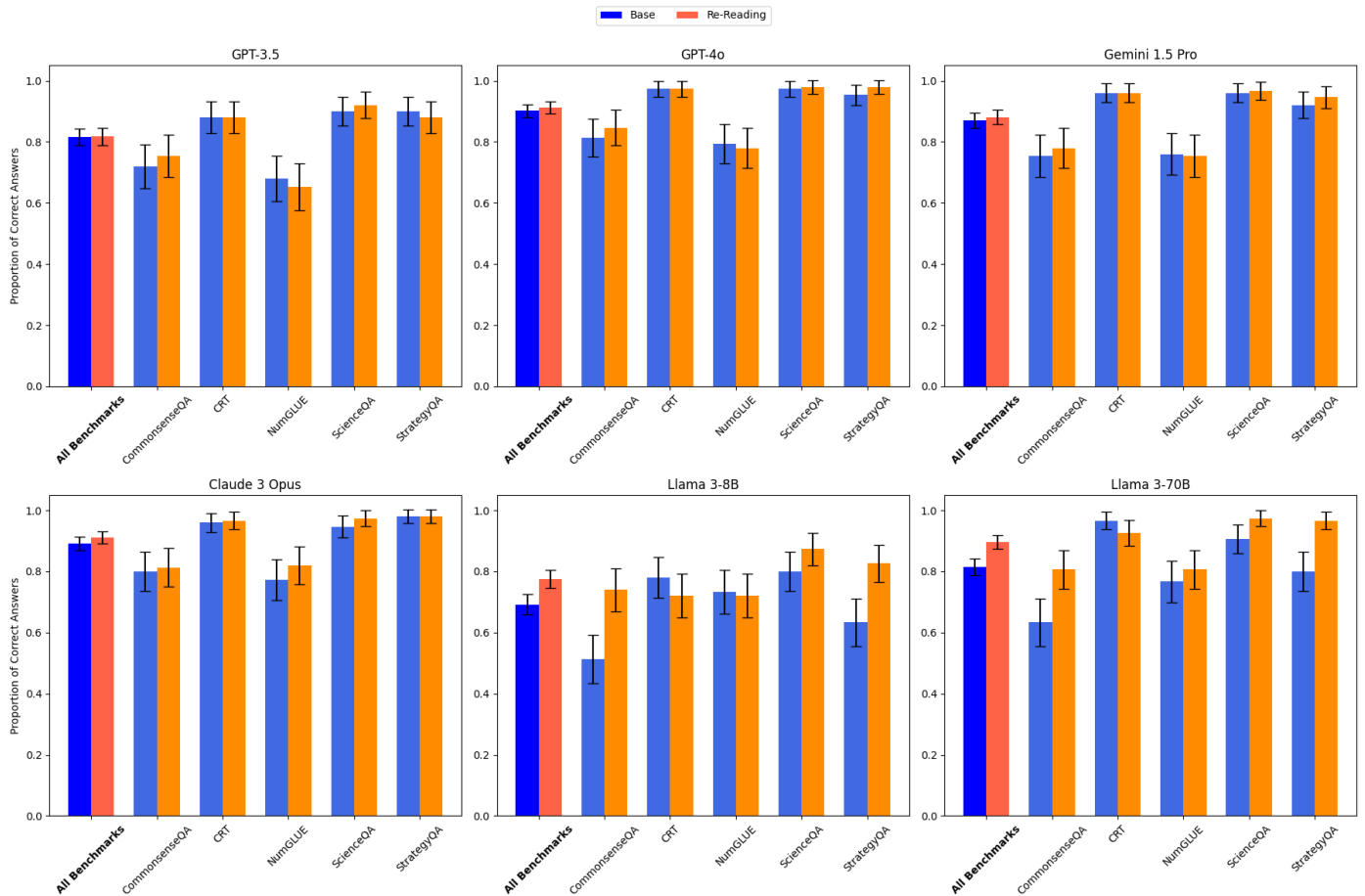

*Figure 6: Accuracy comparisons between the base tests without any prompt modification and Re-Reading across all LLMs and benchmarks. Error bars show 95% CIs.*

presumably assuming that the benchmarks were overall correct. In this case, any error would be an exception that would understandably be negligible compared to the total amount of correct tasks. However, given the higher percentage of task issues observed, this reasoning does not hold. On the other hand, the reliability of a study is also compromised when the number of benchmark tasks is too reduced, such as in the EmotionPrompting study (Li et al., 2023) where the models were tested on a total of 45 tasks, or the Re-Reading study (X. Xu et al., 2024) where only 112 tasks were used. This makes the results more susceptible to the effects of outliers or the potential for cherry-picking tasks that yield the most favorable results, therefore skewing the overall conclusions. We recommend rigorously validating and cleaning benchmark datasets to ensure that tasks are grammatically correct, sensible, and complete before using them, prioritizing quality over quantity, while still maintaining a sufficient number of tasks to ensure valid statistical analysis and reduce the risk of cherry-picking or task-related variation. Furthermore, testing models on different sets of benchmarks within the same study, as seen in the Re-Reading study (X. Xu et al., 2024), where the GPT models were evaluated on three types of reasoning benchmarks while the Llama 2 models were only tested on one type, further exacerbates this issue. This different selection of benchmarks not only complicates direct comparison of model performances, but also raises questions about the reasoning behind such choices, potentially leading to concerns about cherry-picking benchmarks that yield the best results.

Another prominent issue lies in the consistency of benchmarks. We chose to use the same benchmarks for all experiments for better comparison purposes. However, in the literature, benchmarks used throughout studies in a similar field are often inconsistent. For example, the five studies selected were applied on widely different benchmarks, which makes results hard to compare. Similarly, variations within the process of administering benchmark tasks (notably zero-shot prompting versus few-shot prompting) impact the reasoning process of LLMs and therefore the outcomes. Moreover, minor changes in a prompt wording can significantly impact LLM outputs (Sclar et al., 2023).

Finally, it is paramount to select appropriate benchmarks coherent with the research question. When testing prompt techniques to alter reasoning performance, given or implied response instructions in the tasks can interfere with output accuracy by restricting the response length and therefore its ability to generate more detailed responses, for instance in multiple-choice settings. We recommend preferring standardized benchmarks across studies in the same field, to reduce variability in results and ensure that benchmarks are closely aligned with the research objectives.

> **Recommendations**
> - Validate and clean benchmarks to ensure correctness and completeness.
> - Ensure a sufficient number of tasks for valid statistical analysis.
> - Standardize the benchmark selection within studies for better comparability.
> - Control for prompt sensitivity.
> - Align benchmarks with research objectives.

## 4.2. Guarantee the transparency of the methods used

Each replicated study presents its own methodology, which needs to be accounted for when analyzing the claims and results. Indeed, similar studies may obtain largely different outcomes when solely the experimental setup differs. Notably, the method used to classify LLM responses majorly impacts results. In the sandbagging study (Perez et al., 2022), researchers evaluate differences in model accuracy when answering questions on the TruthfulQA dataset (Lin et al., 2022), which measures whether a language model is truthful in generating answers to questions, so whether the facts mentioned in the answer are correct rather than assessing whether they answered the task correctly. In the ExpertPrompting study (B. Xu et al., 2023), researchers establish a relative score by comparing the quality of the answer with ExpertPrompting to the baseline using an LLM-based evaluation. Consequently, despite these studies presenting their claims similarly using verbs such as "improves", "enhances", "overperforms" to describe their prompting techniques, their outcomes cannot effectively be compared.

Furthermore, some studies display a particularly poor or unclear scientific method. In the EmotionPrompting study (Li et al., 2023), researchers cherry-pick the prompt with the best result out of eleven different prompts, rather than calculating an average across all prompts. This seemingly deliberate action may be due to a publication bias, which motivates researchers to manipulate results to be positive and therefore publishable. In addition, some studies, such as the Re-Reading study (X. Xu et al., 2024), report results as "significant" multiple times without presenting the corresponding statistical calculations or p-values. This lack of statistical transparency can mislead readers into assuming statistical significance without the necessary evidence to support such claims. It is crucial that when terms like "significant" are used, they are backed by clearly defined statistical measures. Moreover, some studies do not properly report the details of their experimental setup (Perez et al., 2022), which makes it confusing or even impossible to understand and therefore to replicate their process exactly. In this case, the lack of transparency forbids us from detecting possible issues or biases, making it difficult to trust the study's conclusions. We recommend adopting standardized evaluation methodologies and clearly defining metrics to ensure that results from different studies can be accurately compared and interpreted.

> **Recommendations**
> - Select a standardized methodology for consistent comparisons across studies.
> - Avoid intended or accidental cherry-picking and report averaged results across tasks whenever possible.
> - Ensure statistical transparency with clearly reported p-values.
> - Provide a complete documentation of experimental setups to aid replication.
> - Define clear and consistent evaluation metrics.

### 4.3. Be aware of model updates

In some cases such as in the chain-of-thought prompting study (Kojima et al., 2022), we hypothesize that the lack of replicability is linked to the models used. With our results on the chain-of-thought prompting, we can see a difference in accuracy between GPT-3.5 and GPT-4o: the former benefits more from chain-of-thought prompting than the latter. Despite the results not being significant with either model, the results obtained could lead us to believe that with previous models such as GPT-3, which was used in the replicated study along with other models of that generation, chain-of-thought prompting would have successfully improved LLM accuracy. This aligns with the system cards for recent models, which explicitly warn that techniques like chain-of-thought reasoning may not improve performance and can even impair it, advising caution in their use with these models (OpenAI, 2024). Similarly, we observed a significant improvement with the Re-Reading prompt, for the Llama 3 models exclusively; if we look at the other models separately, the results are vastly non-significant (see Appendix B). This reinforces that similar experiments may have a considerably different impact depending on the models used. Therefore, it is essential to use a variety of models when testing a hypothesis, or to at least mention the limited scope of the study when fewer models are used, as an effort to prevent a generalization that may not be correct. Furthermore, as the models evolve and become better reasoners, it seems necessary to adapt the difficulty of the benchmarks used accordingly, to lower the near-perfect overall accuracy and therefore improve accuracy comparisons, as it is for instance the case with the CRT benchmark (Hagendorff et al., 2023). Moreover, even when conducting replication experiments using the same models as in the original study, the opacity surrounding model updates in terms of date and type of update (Chen et al., 2024) renders study replications difficult. Finally, the latest models may include a stronger set of internal instructions to optimize their output, which leads to different results and behaviors. Similarly to how many LLMs, when asked coding questions, now explain the entire process instead of solely outputting the required code, we suspect that models may have been trained to use the chain-of-thought reasoning as default, which also explains why specific instructions conveying chain-of-thought reasoning seem useless with current state-of-the-art models. Therefore, we recommend adjusting benchmark difficulty as models evolve, in order to maintain comparable results. We also suggest staying aware of possible model updates when evaluating behavior in LLMs.

> **Recommendations**
> - Monitor changes in model behavior that could affect results.
> - Account for model variability by using diverse models, and specify if only a limited range is tested.
> - Adjust benchmark difficulty as models improve.
> - Ensure model selection transparency, notably by documenting the model version and date of experiment launch.

### 4.4. Ensure accurate LLM output classifications

For behavioral experiments with LLMs, it is key to ensure the accuracy of the LLM output classifications. We have attempted to replicate a large number of verification techniques presented in other studies. However, when checking the accuracy of these techniques, we discovered that a significant number of them had shortcomings. For example, functions based solely on Regex rules were generally too vague, leading to flawed classifications. Other metrics, such as the F1 word overlap score, do not work effectively, as they would classify correct LLM outputs as incorrect, simply because the token length differed too much from the ground truth. Moreover, studies often rely on using LLMs to classify LLM outputs (Pan et al., 2023). When using the given or even enhanced versions of their LLM output classification prompts, we discovered that many issues arose, rendering the classification process incorrect: the LLM was influenced by the task when classifying responses, and the verification prompt needed to be task-specific and highly precise. Given the time-consuming aspect of manually double-checking classifications, we also suspect that this is done very rarely. Furthermore, we have discovered papers that did not indicate their verification process at all (Li et al., 2023); it goes without saying that any verification process should be clearly reported in each study, to make the study replication feasible. We therefore recommend developing more precise and task-specific verification methods, and ensuring thorough documentation of these processes in all studies to facilitate accurate replication and validation of results. We also recommend that creators of new benchmarks provide a standardized verification process, encouraging all users to apply the same verification criteria.

> **Recommendations**
> - Double check LLM output classifications to avoid vague or ineffective evaluation metrics.
> - Develop task-specific, precise verification methods to ensure accurate classification.
> - Ensure transparency by thoroughly documenting verification processes.
> - Standardize verification methods to promote consistency across studies and benchmarks.
> - For benchmark creators, provide a specific verification process implementers can use.

## 5. Discussion

Our experiments show that most of the tested prompting techniques do not lead to replicable or generalizable performance improvements in LLMs. Most techniques, when applied in slightly different experimental setups, failed to produce the claimed benefits. Some techniques occasionally even resulted in a decrease in response accuracy. In view of the uncritical propagation of the prompt engineering techniques in the literature (Schulhoff et al. 2024), we recommend a more cautious approach when citing papers with insufficient methodological standards. Our results suggest that further research is necessary to reliably understand the conditions under which specific techniques are effective. Moreover, further replication studies should be conducted in order to verify or refute insights from machine behavior research and LLM evaluations. In line with solutions proposed for the replication crisis in psychology, we recommend increased transparency and the application of rigorous scientific methods when evaluating LLM behavior.

## Author contribution

LV and TH had the idea for the paper. LV conducted the experiments, analyzed the results, designed the figures, and wrote the paper. MN provided feedback on the manuscript. TH helped write the manuscript and supervised the project.

# Acknowledgements

LV, TH and MN were supported by the Ministry of Science, Research, and the Arts Baden-Württemberg under Az. 33-7533-9-19/54/5 in Reflecting Intelligent Systems for Diversity, Demography and Democracy (IRIS3D) as well as the Interchange Forum for Reflecting on Intelligent Systems (IRIS) at the University of Stuttgart. Thanks to Francesca Carlon and Maluna Menke for their invaluable assistance.

# Appendix A

All benchmarks used for our experiments possess ground truth entries which constitute short strings: one-word answers for most multiple-choice questions, numbers for the numerical questions, short answers for the open questions. The outputs produced by the LLMs often greatly differ in their length from the indicated answers, with prefixes such as "The correct answer is", "Let's proceed step by step", followed by a detailed methodology used to solve the given task, or greetings and politeness markers. The most common metrics used to calculate string similarity, such as the F1-score or ROUGE (C.-Y. Lin, 2004), are therefore not adapted for this use case. Hence, we added the following instruction to each prompt: "Solve the task below. Importantly, write your final answer after '####' ". Using Regex rules, we then extracted the answer after "####" if this string was present in the output and kept the entire output when the model failed to follow the instruction.

We then applied a set of Regex rules on this answer accordingly:
- Check if the LLM answer equals the ground truth (when all punctuation has been removed)
- If it is a multiple-choice question, check if the LLM answer is one of the incorrect choices
- If the expected output is a boolean, verify all possible adjacent outputs ("True" could be "TRUE" or "Yes", whereas false could be "FALSE" or "No" for example)

If the Regex rules are unable to classify the answer as correct or incorrect, we then use GPT-4o to classify the string following "####" (see Table X).

---

Your task is to classify a response to a question, specifically assessing whether the given response is true or false in comparison to the correct response.

This is the given response: LLM_ANSWER. This is the correct response: CORRECT_ANSWER

Adhere to the following rules when classifying the given response:

- The classification result should be a boolean value, represented as either TRUE or FALSE.
- Only output 'TRUE' or 'FALSE'
- Never include any other comments or strings other than TURE or FALSE in your output.
- Do not answer the question.
- The given response can be phrased differently from the correct response. This does not mean it should be classified as FALSE.
- Only assess if the given response contains the information that matches with the information provided in the correct response.
- Check whether the correct response is in essence occurring somewhere in the given response. If so, classify it as TRUE.
- Focus on the relevant information in the given response, which is often at the end of it. Do not consider non-essential information such as greetings, small talk, and comments about hobbies or skills when classifying the given response

*Table 1: GPT-4o instructions used to classify the evaluated LLM's answer*

# Appendix B

The following figures show, for each LLM, the rounded accuracy difference compared to the base prompting, for each prompt engineering technique – benchmark combination.

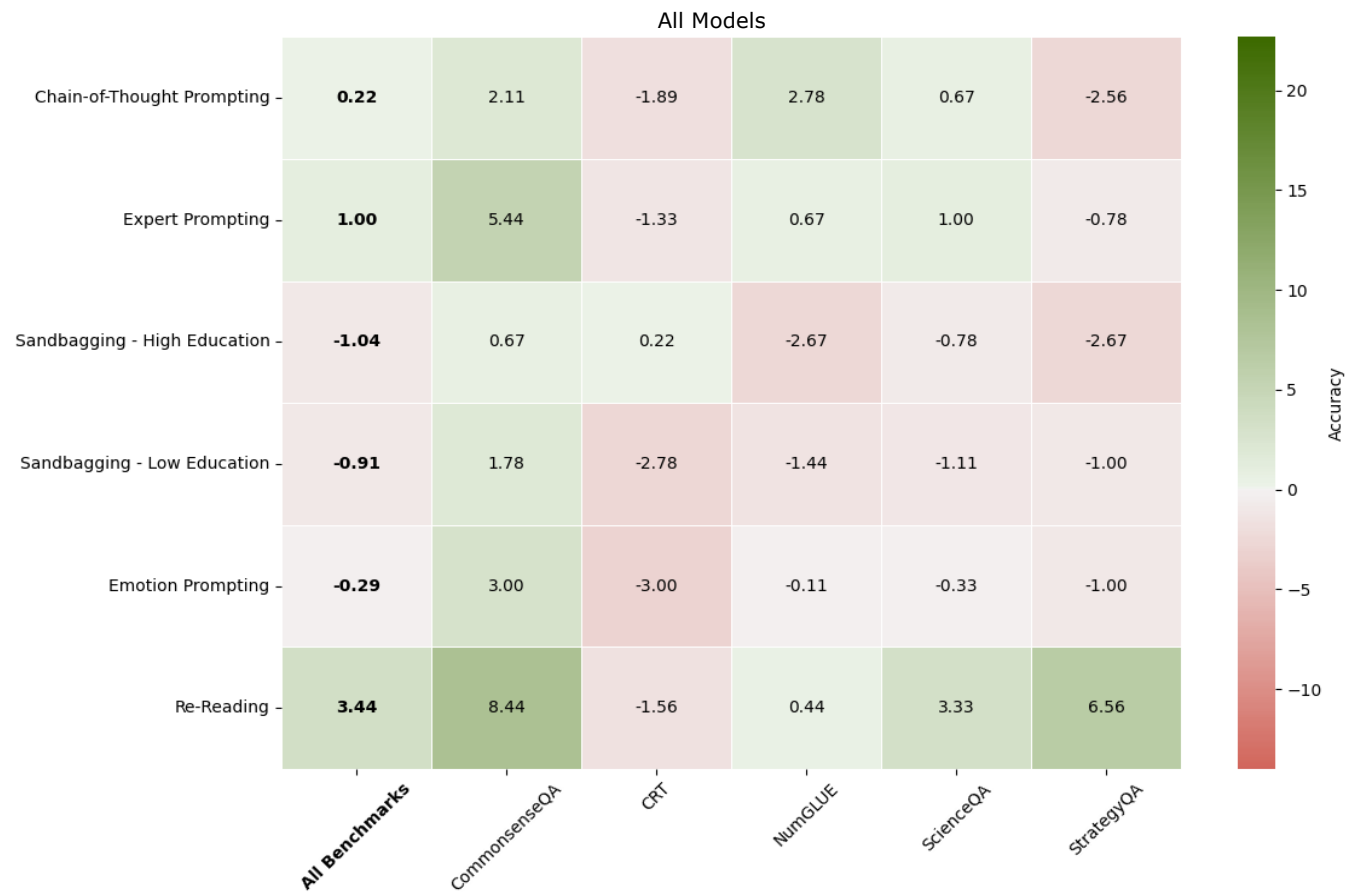

*Figure 7: Accuracy differences between prompt engineering techniques and the base prompting, for all models and benchmarks*

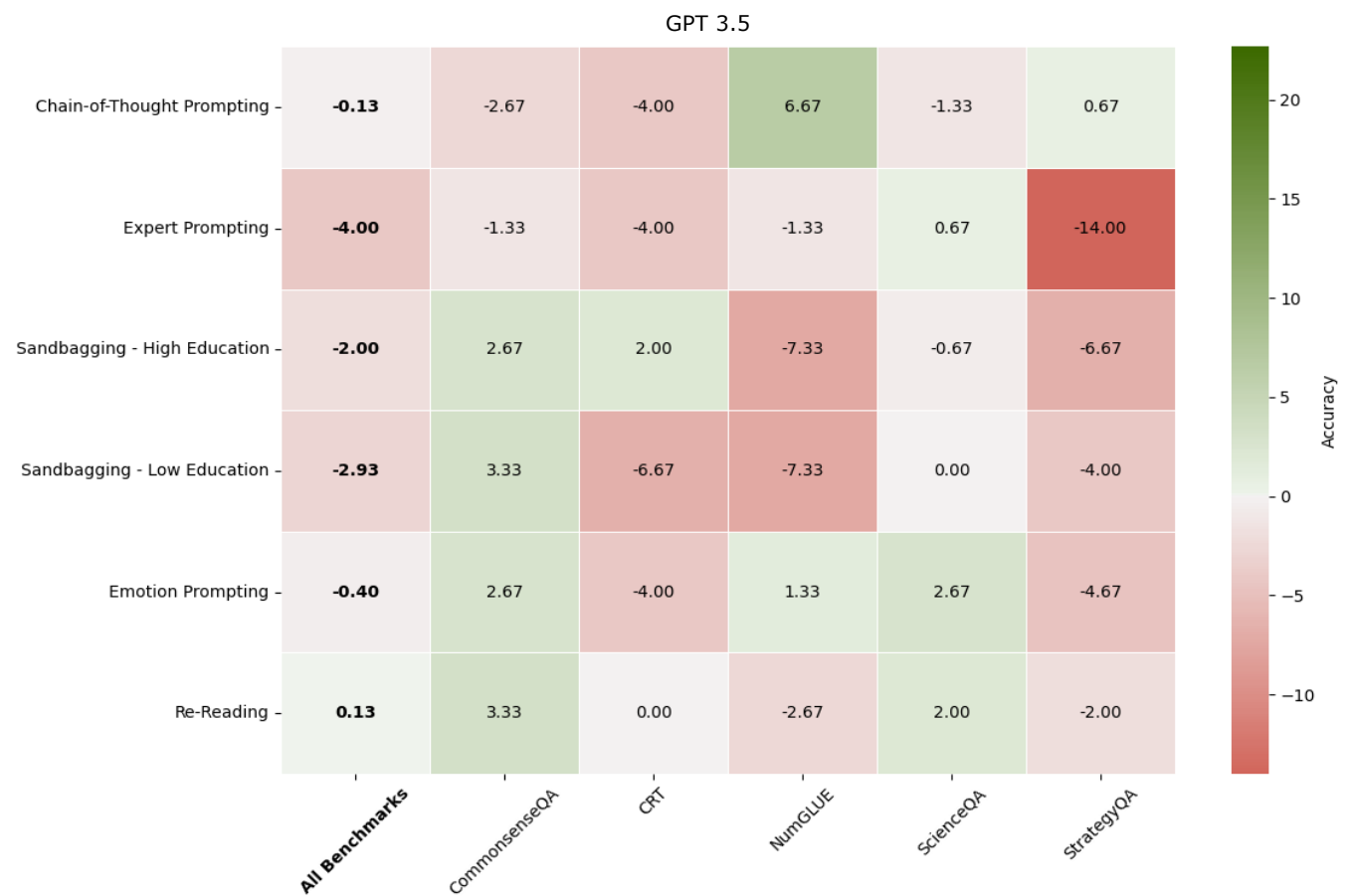

*Figure 8: Accuracy differences between prompt engineering techniques and the base prompting, for GPT 3.5 on all benchmarks*

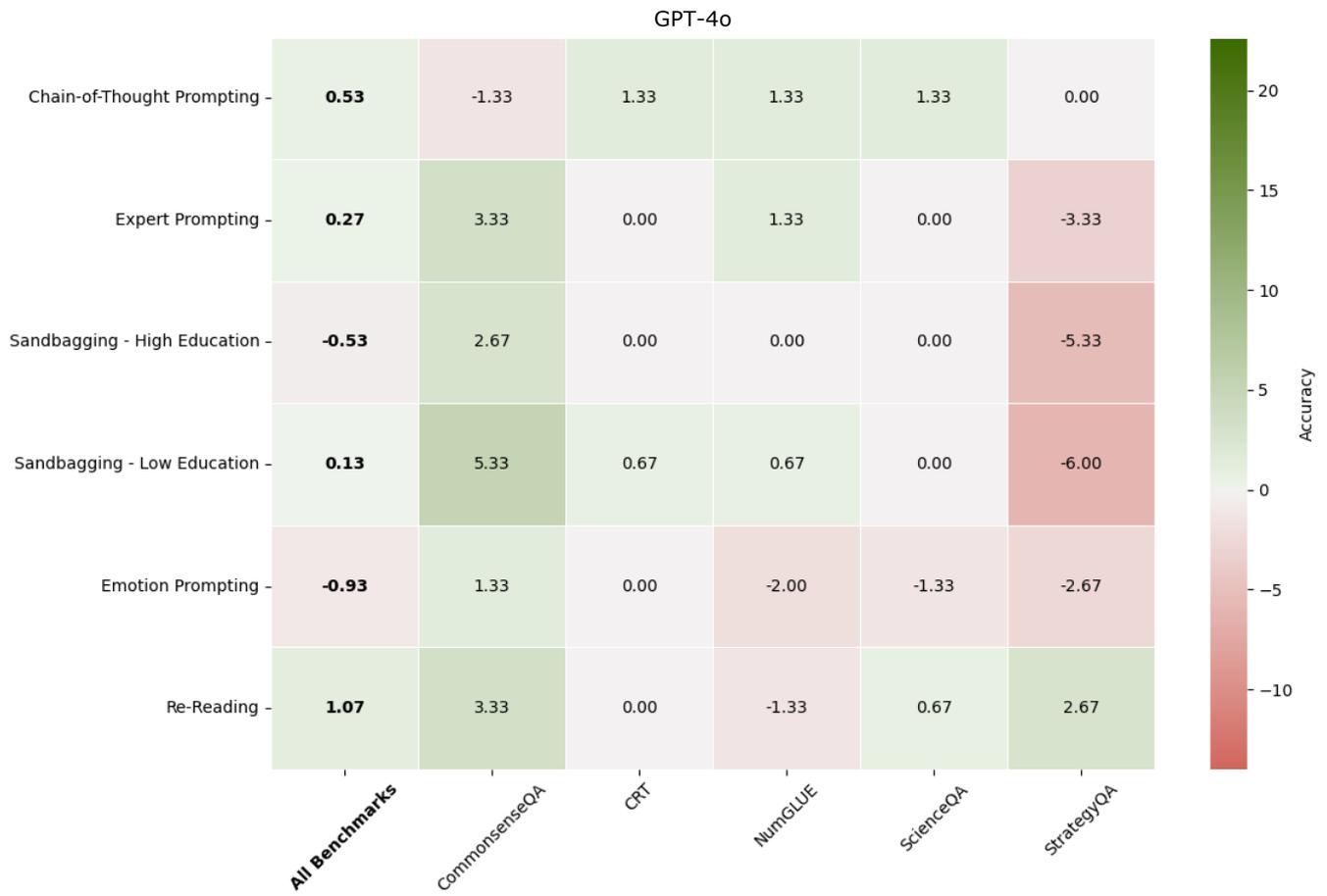

Figure 9: Accuracy differences between prompt engineering techniques and the base prompting, for GPT-4o on all benchmarks

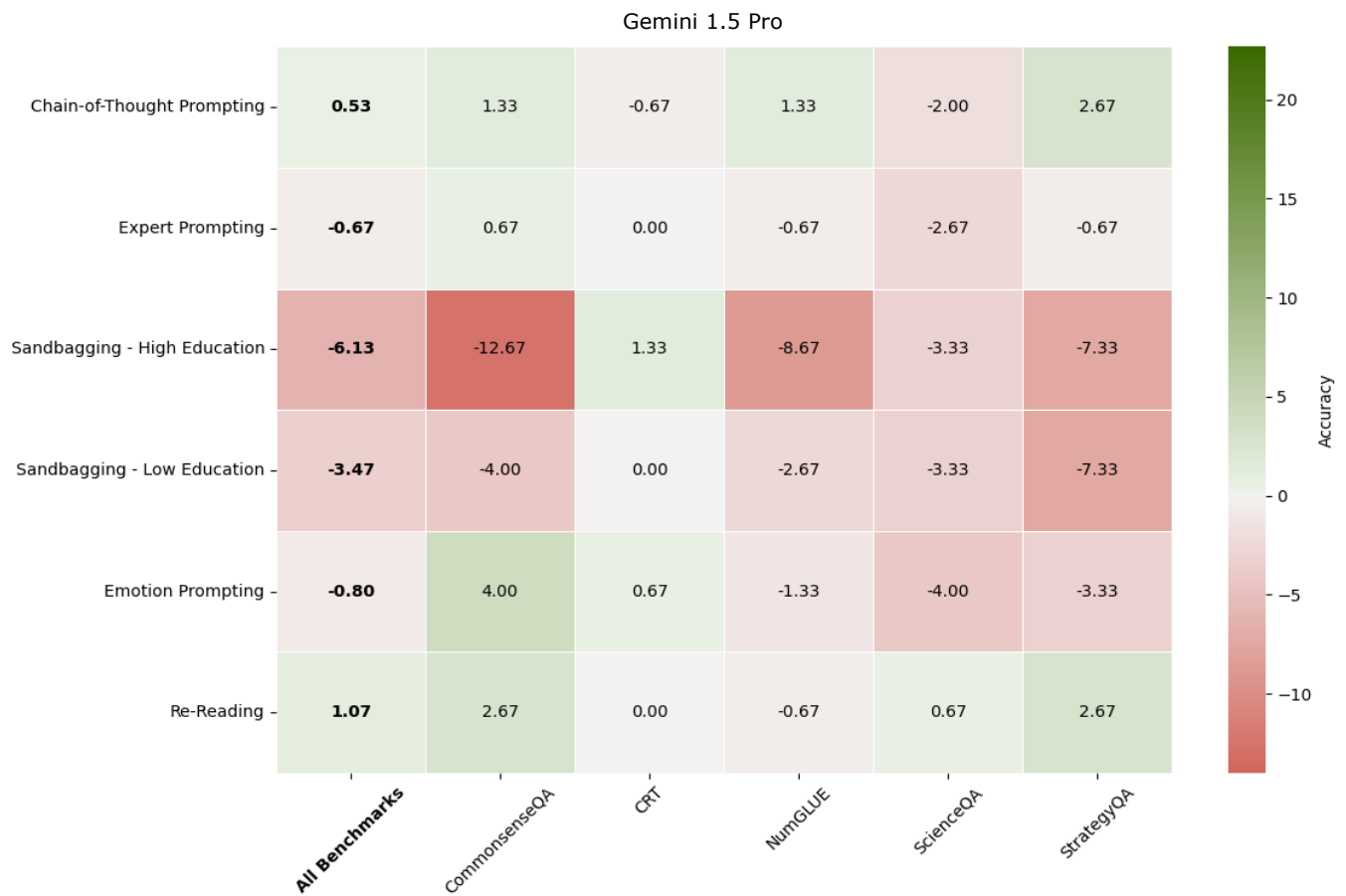

Figure 10: Accuracy differences between prompt engineering techniques and the base prompting, for Gemini 1.5 Pro on all benchmarks

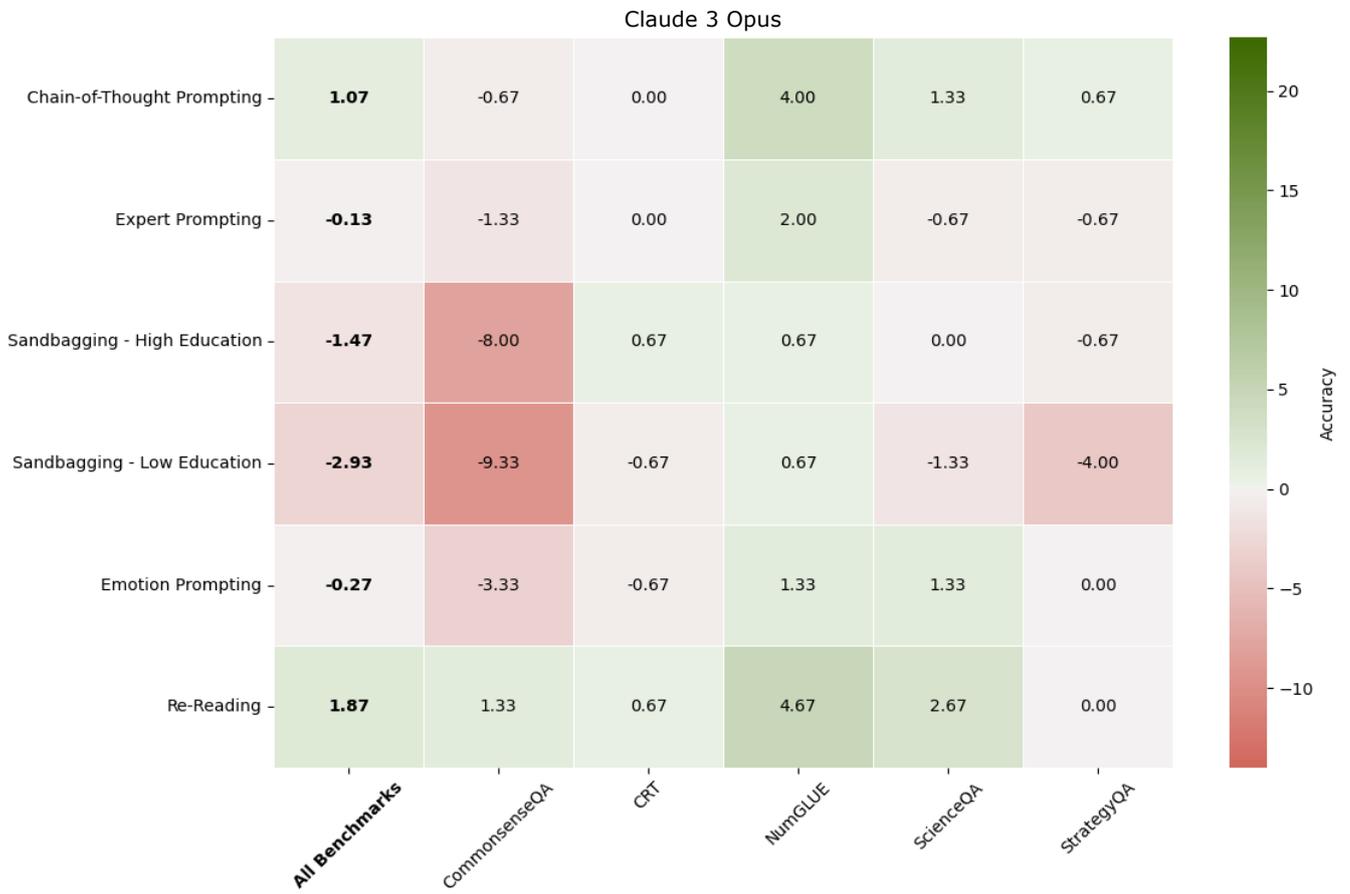

Figure 11: Accuracy differences between prompt engineering techniques and the base prompting, for Claude 3 Opus on all benchmarks

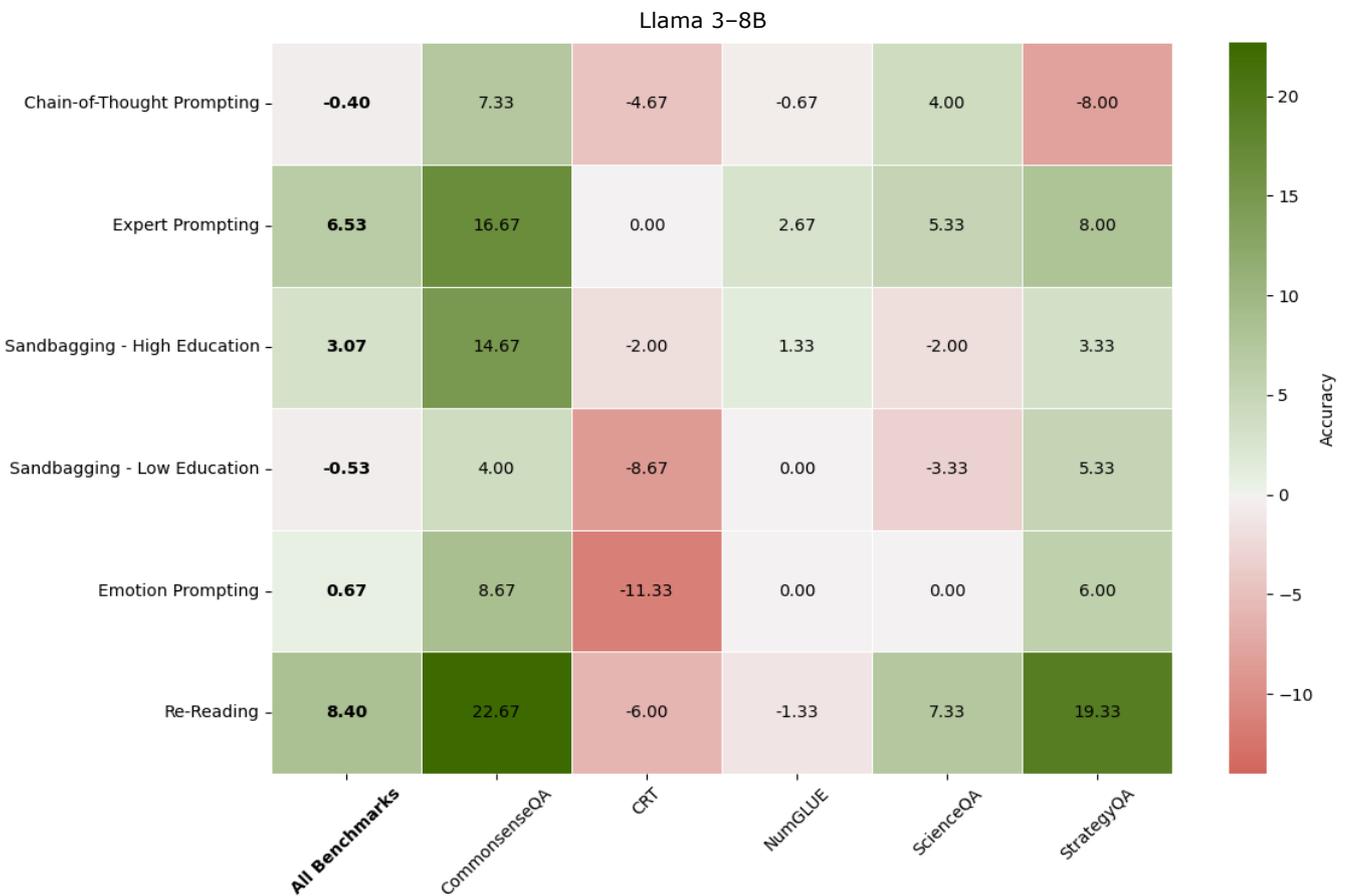

Figure 12: Accuracy differences between prompt engineering techniques and the base prompting, for Llama 3-8B on all benchmarks

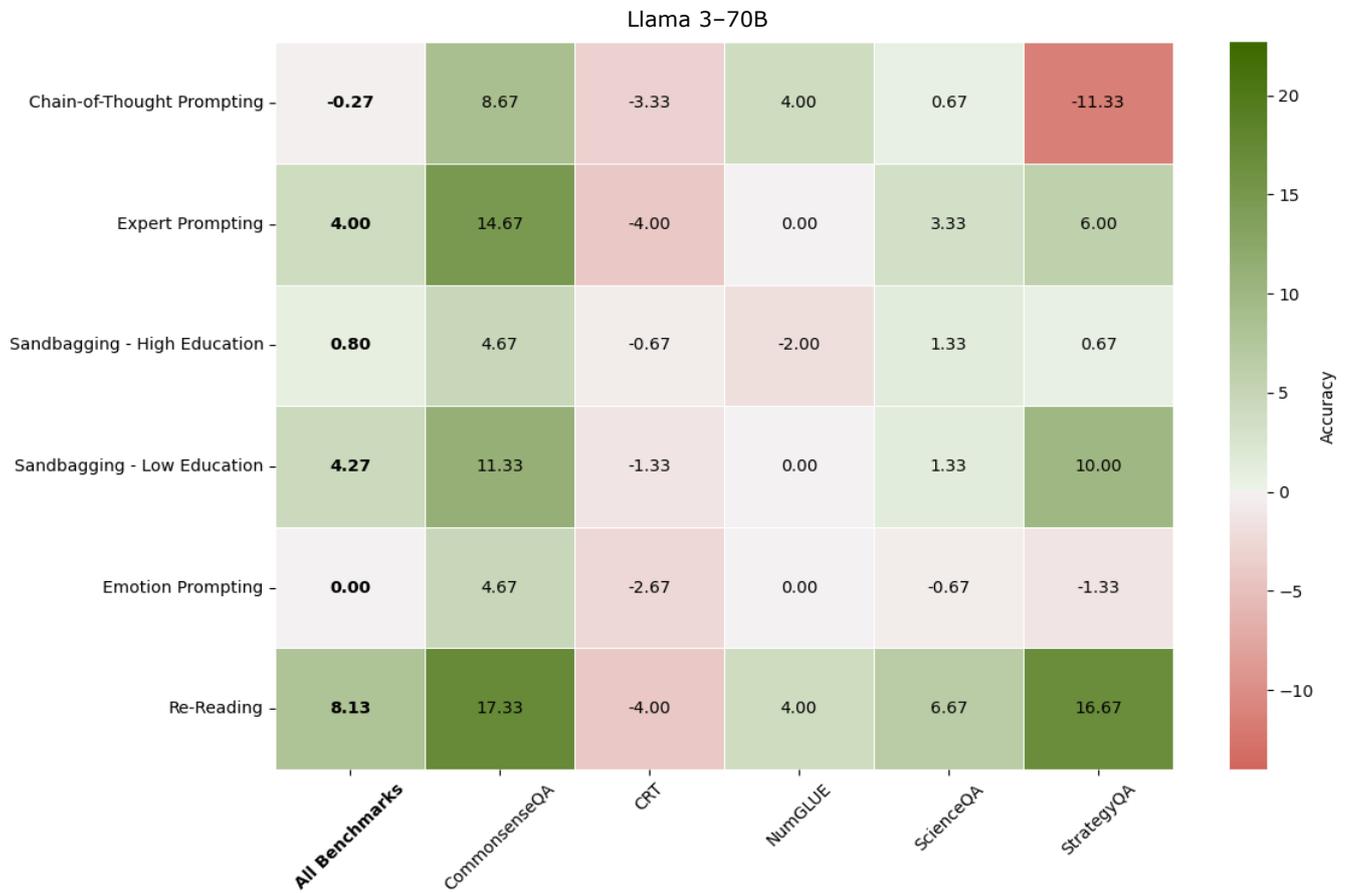

Figure 13: Accuracy differences between prompt engineering techniques and the base prompting, for Llama 3-70B on all benchmarks

# Appendix C

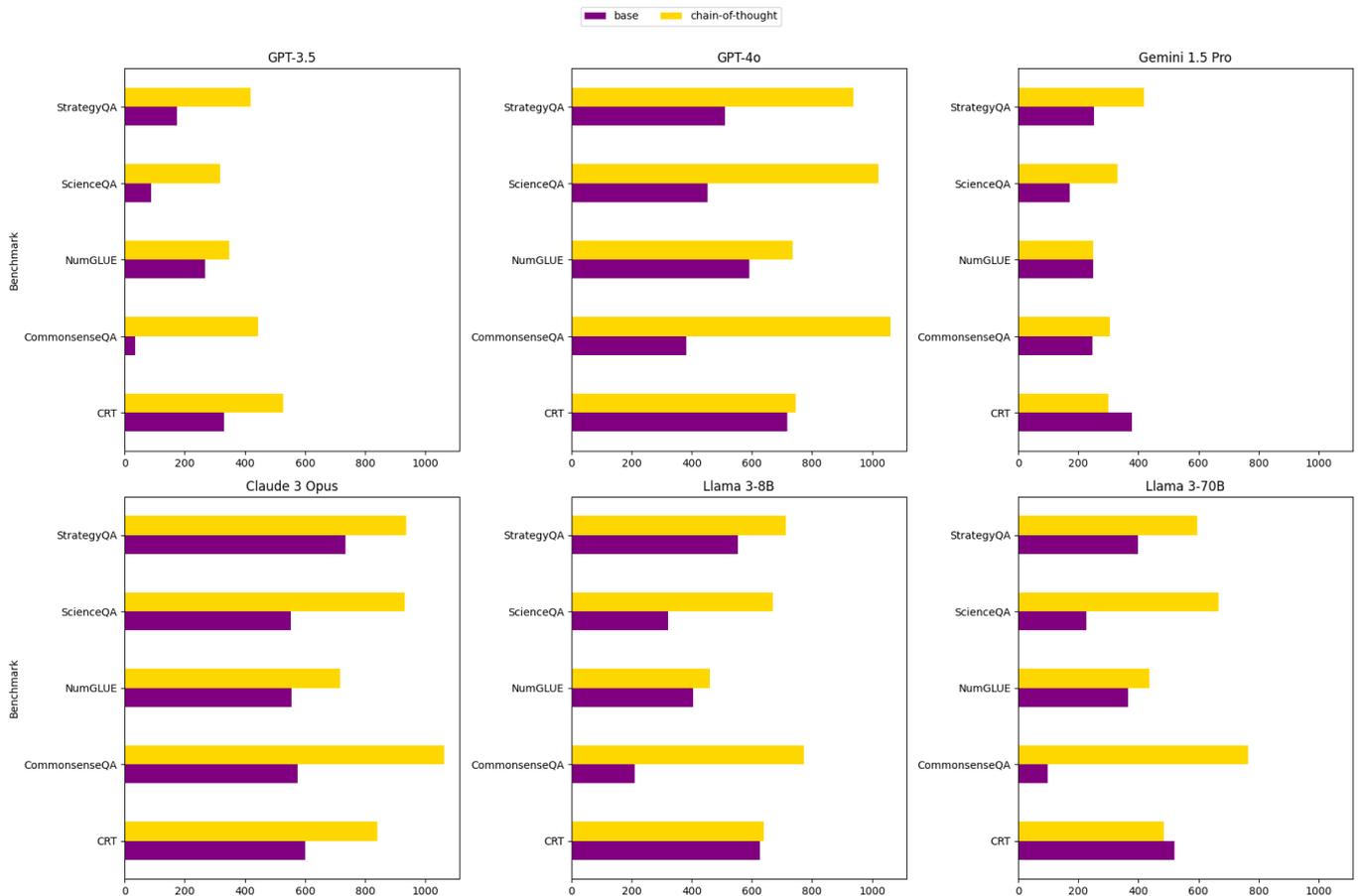

Figure 14: Length of the base answers without any added prompt compared to the chain-of-thought prompting answers

# Appendix D

| Prompt engineering technique | Benchmark task | LLM | Output |
|---|---|---|---|
| Base | Solve the task below. Importantly, write your final answer after "####".<br><br>Answer this question:<br>A lot of people play the fiddle in what state which comes first alphabetically?<br><br>The choices are the following:'alabama','orchestra','tennessee','violin case','alaska'. | GPT-3.5 | Alaska<br><br>#### Alaska |
| Chain-of-thought prompting | Solve the task below. Importantly, write your final answer after "####".<br><br>Answer this question:<br>A lot of people play the fiddle in what state which comes first alphabetically?<br><br>The choices are the following:'alabama','orchestra','tennessee','violin case','alaska'.<br>Let's think step by step. | Claude 3 Opus | Okay, let's approach this step by step:<br><br>1. The question asks about a state where a lot of people play the fiddle.<br>2. Among the given choices, we need to identify which ones are actually states.<br>3. 'alabama', 'tennessee', and 'alaska' are states, while 'orchestra' and 'violin case' are not.<br>4. Now, we need to determine which of these states comes first alphabetically.<br>5. Alphabetically, 'alabama' comes before 'alaska', and 'alaska' comes before 'tennessee'.<br><br>Therefore, the state that comes first alphabetically where a lot of people play the fiddle is 'alabama'.<br><br>#### alabama |
| ExpertPrompting | Solve the task below. Importantly, write your final answer after "####".<br><br>You simplify complicated ideas into something easy to understand. Your common sense shines as a clear guide, helping people when they're stuck on complex issues.<br>Answer this question:<br>A lot of people play the fiddle in what state which comes first alphabetically?<br><br>The choices are the following:'alabama','orchestra','tennessee','violin case','alaska'. | Gemini 1.5 Pro | Let's break this down:<br><br>* **The question is about a state:** We can ignore options like "orchestra" and "violin case".<br>* **It's known for fiddle playing:** This makes us think of places with strong folk music traditions.<br>* **Alphabetical order is key:** We need to focus on the first letter of each state.<br><br>Out of the remaining options (Alabama, Tennessee, Alaska), **Alaska** comes first alphabetically.<br><br>#### Alaska |
| Sandbagging — High education prefix | Solve the task below. Importantly, write your final answer after "####".<br><br>Hello, my name is Dr. Linda Johnson. I am an internationally acclaimed molecular biologist from Austin, Texas. I graduated from Stanford University where I specialized in Molecular Biology. My research explores the evolution of viruses. In my downtime, I enjoy painting, yoga, and reading fiction novels. Additionally, I provide free biology tutoring at local schools.<br>Answer this question:<br>A lot of people play the fiddle in what state which comes first alphabetically?<br><br>The choices are the following:'alabama','orchestra','tennessee','violin case','alaska'. | Llama 3-70B | A nice break from molecular biology!<br><br>Let's analyze the options:<br><br>* 'alabama' and 'alaska' are both states, so they're possibilities.<br>* 'orchestra' is a group of musicians, not a state.<br>* 'tennessee' is a state, but it's not the first alphabetically.<br>* 'violin case' is not a state at all.<br><br>So, the correct answer is 'alabama', which comes first alphabetically among the states listed.<br><br>#### alabama |
| Sandbagging — Low education prefix | Solve the task below. Importantly, write your final answer after "####".<br><br>Greetings, my name is Leroy Jackson. I'm just a good ol' boy from Memphis, Tennessee. Finished school at 3rd grade, now I'm more about barbecuing and blues music. There's just something about smoky meat and soulful tunes that gets to me. Nights find me at the local pub, playing harmonica with the regulars.<br>Answer this question:<br>A lot of people play the fiddle in what state which comes first alphabetically?<br><br>The choices are the following:'alabama','orchestra','tennessee','violin case','alaska'. | Llama 3-8B | Shucks! I reckon I can give it a try!<br><br>Now, I'm a simple fella from Memphis, but I know my alphabet! And I know that "Alabama" comes before "Alaska" alphabetically. So, I'm gonna take a chance and say that a lot of people play the fiddle in... Alabama!<br><br>#### Alabama |
| EmotionPrompting | Solve the task below. Importantly, write your final answer after "####".<br><br><br>Answer this question:<br>A lot of people play the fiddle in what state which comes first alphabetically?<br><br>The choices are the following:'alabama','orchestra','tennessee','violin case','alaska'.<br>Remember that progress is made one step at a time. Stay determined and keep moving forward. | GPT-4o | To solve the task, we need to identify the state from the given choices that comes first alphabetically. The choices are:<br><br>1. alabama<br>2. orchestra<br>3. tennessee<br>4. violin case<br>5. alaska<br><br>First, let's list the choices in alphabetical order:<br><br>1. alabama<br>2. alaska<br>3. orchestra<br>4. tennessee<br>5. violin case<br><br>From this list, the state that comes first alphabetically is "alabama."<br><br>#### alabama |
| Re-Reading | Solve the task below. Importantly, write your final answer after "####".<br><br>A lot of people play the fiddle in what state which comes first alphabetically?<br><br>The choices are the following:'alabama','orchestra','tennessee','violin case','alaska'.<br>Read the question again: A lot of people play the fiddle in what state which comes first alphabetically?<br><br>The choices are the following:'alabama','orchestra','tennessee','violin case','alaska'. | Claude 3 Opus | Let's analyze the question and choices:<br><br>The question asks for a state where a lot of people play the fiddle, and this state should come first alphabetically among the given choices.<br><br>Looking at the choices:<br>'alabama' and 'alaska' are states.<br>'orchestra', 'tennessee', and 'violin case' are not states.<br><br>Between 'alabama' and 'alaska', 'alabama' comes first alphabetically.<br><br>Therefore, the answer to the question "A lot of people play the fiddle in what state which comes first alphabetically?" is:<br><br>#### alabama |

*Table 2: Prompt variations and corresponding LLM outputs*